%% file: ms.tex
\documentclass[runningheads]{llncs}
\usepackage[pdftex]{graphicx}
\usepackage{subfig}
\usepackage{amsmath,amssymb} 
\usepackage{color}
\usepackage{fourier} 
\usepackage{array}
\usepackage{makecell}
\usepackage{multirow}

\newcommand{\argmin}{\mathop{\rm arg~min}\limits}

\usepackage{xspace}
\usepackage{hyperref}
\usepackage{bm}

\makeatletter
\DeclareRobustCommand\onedot{\futurelet\@let@token\@onedot}
\def\@onedot{\ifx\@let@token.\else.\null\fi\xspace}

\def\eg{\emph{e.g}\onedot} 
\def\ie{\emph{i.e}\onedot}

\def\etal{\emph{et al}\onedot}
\makeatother

\newcommand{\modelname}{SCGAN}

\begin{document}

\title{Shape-conditioned Image Generation by Learning
Latent Appearance Representation from Unpaired Data\thanks{
Yusuke Sugano is supported by and JST CREST Grant Number JPMJCR1781, and Yasuyuki Matsushita is supported by the New Energy and Industrial Technology Development Organization (NEDO), Japan.	
}
} 
\titlerunning{Shape-conditioned Image Generation} 


\author{Yutaro Miyauchi\inst{1} \orcidID{0000-0002-5951-3172} \and
Yusuke Sugano\inst{1}\orcidID{0000-0003-4206-710X} \and \\
Yasuyuki Matsushita\inst{1}\orcidID{0000-0002-1935-4752}}
%

\authorrunning{Y. Miyauchi et al.} 


\institute{
Graduate School of Information Science and Technology, Osaka University, Japan\\ \email{\{miyauchi.yutaro,sugano,yasumat\}@ist.osaka-u.ac.jp}
}

\maketitle

\input{abstract}

\input{introduction}

\input{related_work}

\input{method}

\input{experiments}

\input{conclusion}

\newpage

\input{ref.bbl}
\end{document}

%% file: abstract.tex

\begin{abstract}

Conditional image generation is effective for diverse tasks including training data synthesis for learning-based computer vision. 
However, despite the recent advances in generative adversarial networks (GANs), it is still a challenging task to generate images with detailed conditioning on object shapes.
Existing methods for conditional image generation use category labels and/or keypoints and are only give limited control over object categories.
In this work, we present \modelname, an architecture to generate images with a desired shape specified by an input normal map.
The shape-conditioned image generation task is achieved by explicitly modeling the image appearance via a latent appearance vector.
The network is trained using unpaired training samples of real images and rendered normal maps.
This approach enables us to generate images of arbitrary object categories with the target shape and diverse image appearances.
We show the effectiveness of our method through both qualitative and quantitative evaluation on training data generation tasks.

\end{abstract}

%% file: introduction.tex

\section{Introduction}
Generating realistic images is a central task in both computer vision and computer graphics.
Despite recent advances in generative adversarial networks (GANs), it is still challenging to fully control how the target object should appear in the output images.
There have been several approaches for conditional image generation which introduce additional conditions to GANs such as class labels~\cite{Odena2016,Tan2017} and keypoints~\cite{Ma2017,Reed2016}.
However, previous approaches still suffer from an inability to control detailed object shapes and lack generalizability to arbitrary object categories.

Training data synthesis is one of the most promising applications of conditional image generation.
Since recognition performance of machine learning-based methods heavily depends on the amount and quality of training images, there is an increasing demand for methods and datasets for training recognition models using synthetic data~\cite{Wood2016,Qiu2016,Ros2016,Hinterstoisser2012}.
However, when synthetic training images are rendered with off-the-shelf computer graphics techniques, the trained estimators still suffer from an appearance gap from actual, often degraded test images.
GANs have also been used to modify synthetic data to more realistic training images, and it has been shown that such data can improve the performance of learned estimators~\cite{Shrivastava2016,Sixt2016,Silberman}.
These methods use synthetic data as a condition on image generation so that output images remain visually similar to the input images and therefore keep their original ground-truth labels.
In this sense, the aforementioned limitation of conditional image generation severely restricts the application of such training data synthesis approaches.
If the method allows for more fine-grained control of object shapes, poses, and appearances, it can open a way for generating training data for, \eg, generic object recognition and pose estimation.

\begin{figure}[t]
	\begin{center}
		\includegraphics[width=\linewidth]{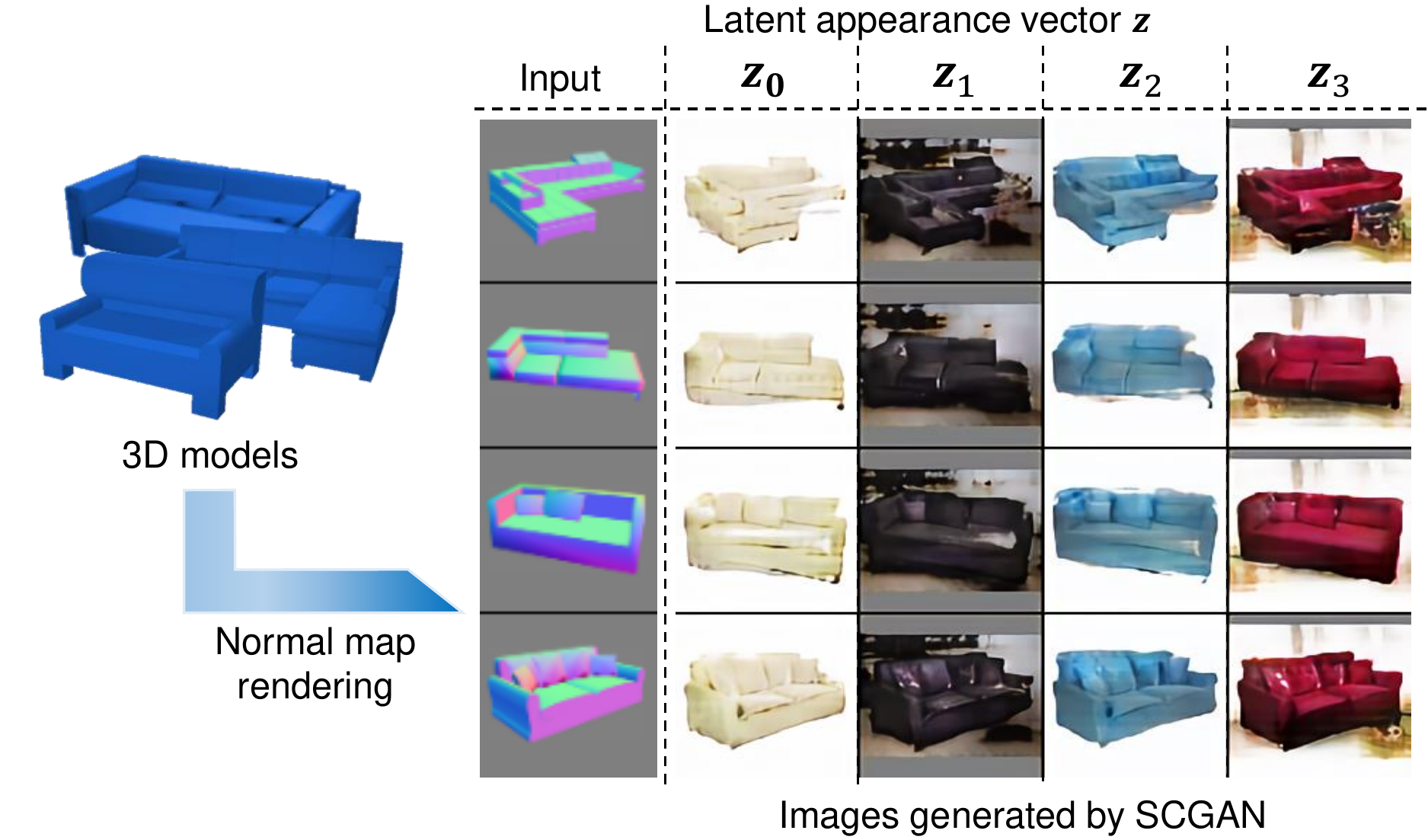}
	\end{center}
	\caption{The proposed shape-conditioned image generation network (\modelname) outputs images of an arbitrary object with the same shape as the input normal map, while controlling the image appearances via latent appearance vectors.}
	\label{fig:Teaser}
\end{figure}

In this work, we propose \modelname~(Shape-Conditioned GAN), a GAN architecture for generating images conditioned by input 3D shapes.
As illustrated in Fig.~\ref{fig:Teaser}, the goal of our method is to provide a way to generate images of arbitrary objects with the same shape as the input normal map.
The image appearance is explicitly modeled as a latent vector, which can be either randomly assigned or extracted from actual images.
Since we cannot always expect paired training data of normal maps and images, the overall network is trained using the cycle consistency loss~\cite{Zhu2017} between the original and back-reconstructed images.
In addition, the proposed architecture employs an extra discriminator network to examine whether the generated appearance vector follows the assumed distribution.
Unlike prior work using a similar idea for feature learning~\cite{Donahue2016}, this appearance discriminator allows us to not only control the image appearance, but also to improve the quality of generated images.
We demonstrate the effectiveness of our method in comparison with baseline approaches through qualitative analysis of generated images, and quantitative evaluation of training data synthesis performance on appearance-based object pose estimation tasks.

Our contributions are twofold.
First, to the best of our knowledge, we present the first GAN architecture which uses normal maps as the input condition for image generation.
This provides a flexible and generic way for generating shape-conditioned images without relying on any assumption on the target object category.
Second, through experiments, we show that the proposed method allows us to generate training data for appearance-based object pose estimation, with better performances than synthetic data generated by baseline GAN architectures.

%% file: related_work.tex

\section{Related work}

Our method aims at generating shape-conditioned images with realistic appearances, related to prior methods on conditional image generation GANs.
One of the potential applications of our method is generating realistic training data, and hence our method is further related to methods applying GANs for bridging the gap between synthetic training data and real images.

\subsection{GANs for Conditional Image Generation}

Generative Adversarial Networks (GANs) have made considerable advances in recent years~\cite{Goodfellow2014,Mao2016,Uality2018,Zhao2016}, and have been successfully applied to various tasks such as image super-resolution~\cite{Ledig2016a,Johnson2016}, inpainting~\cite{Iizuka2017}, and face aging~\cite{Zhang2017b}.
GANs consist of mainly two networks, generator and discriminator, which are trained in an adversarial manner.
The generator generates images so that they are recognized as real ones, while the discriminator learns to discriminate generated images from real images from a training dataset.
The generator usually receives a vector of random numbers sampled from an arbitrary probability distribution as input, and outputs an image through the network.
However, as discussed earlier, most of the standard GAN architectures do not allow for fine-grained control of the output images.

To address this limitation, much research has been conducted on GAN architectures for conditional image generation.
There have been several approaches to use class labels as a condition on generated images and to specify which object category to be drawn in the output image~\cite{Odena2016}.
Similarly, some prior work proposed to control the generated images by conditioning them on human-interpretable feature vectors built in an unsupervised manner~\cite{Chen2016a,Springenberg2016}.
To increase the flexibility of image generation, some works further used input features indicating where and how the target object should be drawn, such as bounding box~\cite{Reed2016} and keypoints~\cite{Ma2017}.
Alternatively, iGAN~\cite{Zhu2016} and the Introspective Adversarial Network~\cite{Brock2016} take an approach to use user drawings as a condition for image generation.
However, the conditions used in these methods still have a limitation that precise 3D shape control is only possible with specific object categories with hand-designed keypoint locations.
In contrast, our method allows for direct control on arbitrary object shapes using normal map rendering, without requiring paired training data.

\subsection{Learning with Simulated/Synthesized Images}

Due to the limited availability of fully-labeled training images for diverse computer vision tasks, there is an increasing attention on synthetic training data.
Computer graphics pipelines have been employed to synthesize images with desired ground-truth labels.
Such a learning-by-synthesis approach is especially efficient for tasks whose ground-truth labels require costly manual annotation, such as semantic segmentation~\cite{Qiu2016,Ros2016} and eye gaze estimation~\cite{Wood2016}.
However, synthetic images still suffer from a large gap from real images in terms of object appearance and often degraded imaging quality, and hence the learned estimator cannot directly achieve desired performance on real-world input images.

To fill the gap between training (synthetic) and test (real) image domains, there have been proposed many domain adaptation techniques.
In addition to research attempts on the learning process~\cite{Zhang2017a,Sun2014,Vazquez2014,Baktashmotlagh2013,Ganin2015}, GANs have been also shown as promising tools for bridging the domain gap.
Shrivastava~\etal~proposed the SimGAN that modifies the input synthetic images to be visually similar to real images, and showed that such an approach improves the baseline performances on tasks like hand pose and gaze estimation~\cite{Shrivastava2016}.
RenderGAN~\cite{Sixt2016} takes a similar approach to convert simple barcode-like input images into realistic images.
CycleGAN~\cite{Zhu2017} architecture provides a way to mutually convert images from two different domains without requiring paired images, and also be applicable to the domain adaptation task.
Bousmalis~\etal~proposed the pixel-level domain adaptation (PixelDA) approach which transfers source images to the target domain under the pixel-level similarity constraint.
Essentially, synthetic images were used as a strong constraint on output images in these methods, and GANs were restricted only to modify the imaging properties of the target object.
In contrast, since our method uses texture-less normal maps to provide purely shape-related information to the generator, it allows for a full flexibility to control object and background appearances.

%% file: method.tex

\section{\modelname: Shape-conditioned Image Generation Network}

The goal of \modelname~is to generate images of arbitrary object categories, with the same shape as the input normal map.
While the training process requires an access to both normal maps and real images of the target object, in practice it is almost impossible to assume paired training data.
To this end, \modelname~adopts the idea of cycle-consistency loss~\cite{Zhu2017} and the whole network is trained using unpaired training images.
Furthermore, to maximize the flexibility of object appearances, the image generator also takes an appearance vector as input, in addition to the normal map.
By training the network so that appearance-related information is represented only with the appearance vector, our method realizes the shape-conditioned image generation task more efficiently and accurately.

\subsection{Network Architecture}

\begin{figure}[t]
	\begin{center}
		\includegraphics[width=\linewidth]{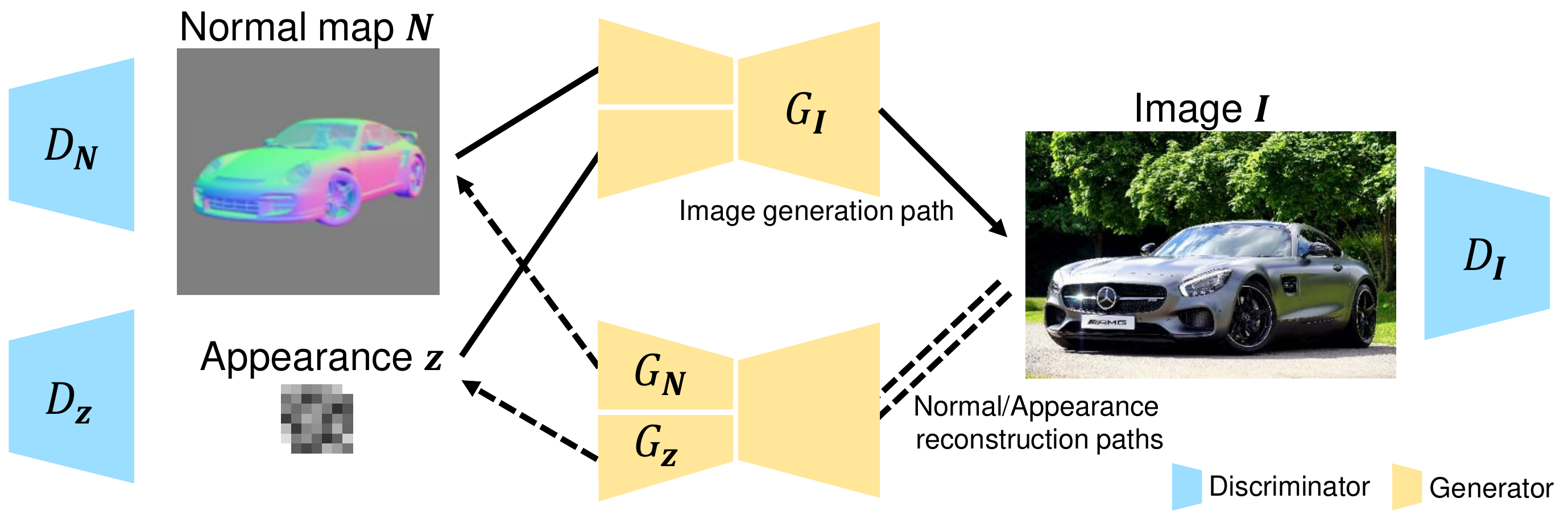}
	\end{center}
	\caption{Overview of the proposed architecture. $G_I, G_N$, and $G_z$ are generators, and $D_I, D_N$, and $D_z$ are discriminators. $G_N$ and $G_z$ share weights at the first few layers, and simultaneously output $\bm N$ and $\bm z$ from the input real image $\bm I$.}
	\label{fig:Architecture}
\end{figure}

As illustrated in Fig.~\ref{fig:Architecture}, the proposed architecture consists of five convolutional neural networks.
$G_I$ is an image generator that takes an appearance vector $\bm{z} \in \mathbb R^n$ and a normal map $\bm{N} \in \mathbb R^{m\times m}$ as input, and outputs an image $\bm{I} \in \mathbb R^{m\times m}$.
Conversely, $G_N$ and $G_z$ correspond to the normal map and appearance vector generators with partially shared network weights that converts an image $\bm{I}$ to an appearance vector $\bm{z}$ and a normal map $\bm{N}$.
Each data modality has their own discriminators $D_I, D_N$, and $D_z$.
While $D_I$ and $D_N$ judge whether the input image and normal map are real or generated, $D_z$ judges whether the input appearance vector is the one sampled from a Gaussian distribution or not.

As described earlier, the proposed network is designed to be trained on unpaired training samples using the cycle-consistency loss~\cite{Zhu2017}.
While the main goal of our approach is to train the image generator $G_I$, normal map and appearance generators ($G_N$ and $G_z$) are also trained and used to back-reconstruct each modality and compare with the original input.
However, if we only consider generators and discriminators of images and normal maps, generators tend to satisfy the cycle-consistency loss by {\em embedding} hidden information to intermediate data.
For example, if the image generator learns to embed input information to the output image, the normal map generator can recover the original normal map without taking into account the object shape in the intermediate image.
To avoid such situations, we also enforce the network to learn to separate shape and appearance information by introducing the appearance generator and discriminator.
The proposed network effectively generates shape-conditioned images by modeling the appearance variation in the training data as a Gaussian appearance vector, while also allowing us to explicitly sample appearance information from actual images using the appearance generator $G_z$.

\subsection{Training Loss}
We train discriminators and generators using the WGAN-GP loss~\cite{Gulrajani2017} which is based on the Wasserstein-1 distance between real and generated data distributions.
The loss functions $L_d$ and $L_g$ for discriminators and generators, respectively, are defined as:
\begin{eqnarray}
L_d(D) &=& \mathbb E_{\bm{x}\sim P_{\bm{x}}} [D(G(\bm{x}))] - \mathbb E_{\bm{\hat{x}}\sim P_{\bm{\hat{x}}}}[D(\bm{\hat{x}})]
		+ \lambda_p \mathbb E_{\bm{\dot{x}}\sim P_{\bm{\dot{x}}}}[(||\nabla_{\bm{\dot{x}}} D(\bm{\dot{x}})||_2 - 1)^2], \label{eq:WGANGPLoss} \\
L_g(G) &=& -\mathbb E_{\bm{x}\sim P_{\bm{x}}} [D(G(\bm{x}))], \nonumber
\end{eqnarray}
where $\bm{x}$ is real data (image, normal map, appearance vector), $\bm{\hat{x}}$ is generated data from their corresponding generators, and $\bm{\dot{x}} = \epsilon G(\bm{x}) + (1-\epsilon)\bm{\hat{x}}. \quad \epsilon \sim \mathcal{U}[0,1]$ is random-weighted sum of input and generated data.
$P_{\bm{x}}$, $P_{\bm{\hat{x}}}$, and $P_{\bm{\dot{x}}}$ indicate distributions of each data, and $\mathbb E$ represents the mean of the distribution.
The third term of Eq.~(\ref{eq:WGANGPLoss}) has an effect of stabilizing the adversarial training~\cite{Gulrajani2017}.

In our implementation, while three discriminators are trained using the individual discriminator losses, all generators are jointly trained as:
\begin{eqnarray}
G_I, G_N, G_z &=& \argmin_{G_I, G_N, G_z} L(G_I, G_N, G_z), \nonumber
\end{eqnarray}
where $L(G_I, G_N, G_z)$ is the joint loss function also taking into account the cycle-consi-stency losses:
\begin{eqnarray}
L(G_I, G_N, G_z) &=&  L_g(G_I) + L_g(G_N) + L_g(G_z) + \lambda_N ||\bm{N} - G_N(G_I(\bm{N},\bm{z}))||^2_F \nonumber \\
	&& + \lambda_I ||\bm{I} - G_I(G_N(\bm{I}), G_z(\bm{I}))||^2_F + \lambda_z ||\bm{z} - G_z(G_I(\bm{N}, \bm{z}))||^2_2. \label{CycleWGANLoss}
\end{eqnarray}
$\lambda_I, \lambda_N$, and $\lambda_z$ are weights for each cycle-consistency loss term which are defined as the distance between the input and the back-reconstructed output.
These weights are required to take balance between discriminator and cycle-consistency losses in each domain, and they control how strictly the model should maintain the input shapes.
Image $\bm{I}$ and normal map $\bm{N}$ are sampled from the distribution of real data $P_I$ and $P_N$, and $\bm{z}$ are an appearance vector sampled from a zero-mean Gaussian distribution $\mathcal{N}(0, \sigma^2)$.

\subsection{Implementation Details}

\begin{figure}[t]
	\begin{center}
		\includegraphics[width=\linewidth]{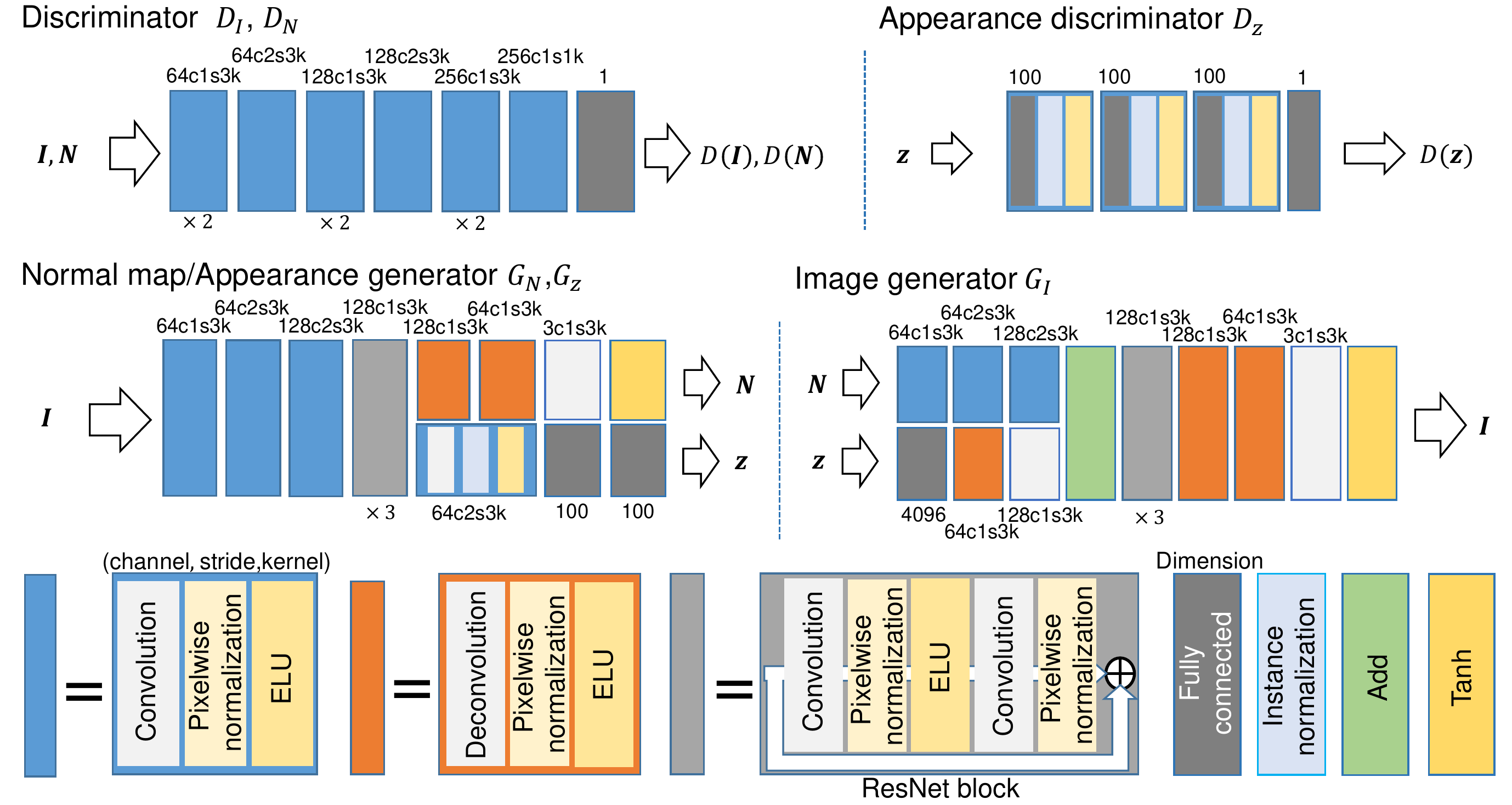}
	\end{center}
	\caption{Details of the generator/discriminator networks. $\bm N,\bm I$, and $\bm z$ indicate normal map, real image, and appearance vector. Parameters of the convolutional layers are indicated as $C$c$S$s$K$k, \ie, a feature map is convolved into $C$ channels with stride $S$ and kernel size $K$.}
	\label{fig:Networks}
\end{figure}

\autoref{fig:Networks} shows the details of generator/discriminator networks.
The architecture of the generator network follows Zhu~\etal~\cite{Zhu2017} and the network mainly consists of convolution ({\em Convolution-Pixelwise normalization-ELU}) block, deconvolution ({\em Deconvolution-Pixelwise normalization-ELU}) block, and ResNet block~\cite{He2016}.
As described earlier, parameters of the first six convolution blocks of the normal map generator $G_N$ and the appearance vector generator $G_z$ are shared.
The discriminator network for images and normal maps also consists of the convolution block and outputs a scalar value indicating discrimination results through a fully connected layer.
The appearance discriminator network consists of a fully connected layer followed by a {\em Instance normalization-ELU} block, and also outputs a scalar value through a fully connected layer.
The size of input images and normal maps are set to $64\times 64$ pixels, and is downsampled to $16\times 16$ before the ResNet blocks.

During training, each discriminator was trained independently with respect to the corresponding discriminator losses.
Then the generators were trained as Eq.~(\ref{CycleWGANLoss}), with the discriminator parameters fixed.
Following \cite{Gulrajani2017}, discriminators were updated five times as much as generators.
The networks are optimized using the Adam algorithm~\cite{Kingma2014} ($\alpha =0.001, \beta_1=0.5, \beta_2=0.9$), with the batch size of $16$.
We fixed hyper-parameters in Eq.~(\ref{CycleWGANLoss}) as $\lambda_I=1.0, \lambda_N=1.0$, and $\lambda_z=10.0$.
The variance of the Gaussian distribution was set to $\sigma^2 = 0.5$.

%% file: experiments.tex

\section{Experiments}

We demonstrate the performance of the proposed \modelname~architecture through both qualitative analysis and quantitative evaluation.
As a qualitative analysis, we compare shape-conditional generated images from the proposed method and other baseline methods in terms of both accuracies of object shape and diversity of object appearances.
In addition, we show some ablation studies to analyze the efficiency of the proposed network design.
As a quantitative evaluation, we further compare the performance of appearance-based object pose estimator using these generated images from different methods as training data.

\subsubsection{Training Datasets}

\begin{figure}[t]
	\begin{center}
	\includegraphics[width=\linewidth]{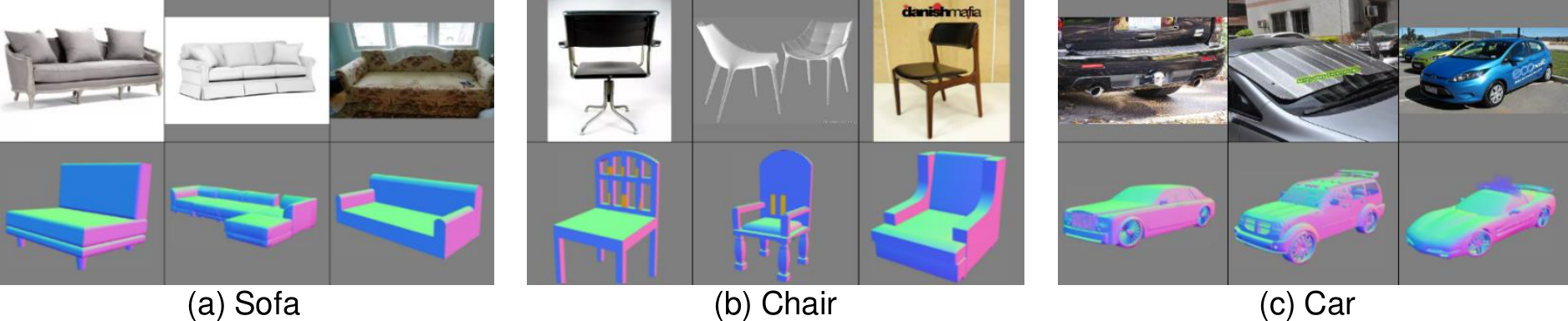}
	\end{center}
	\caption{Examples of the training data taken from the LSUN dataset~\cite{Yu2015} and the ShapeNet dataset~\cite{Chang2015}. The top row is real images from the LSUN dataset after post-processing, and the bottom row is normal maps rendered using models from the ShapeNet dataset.}
	\label{fig:TrainingSamples}
\end{figure}

In both qualitative and quantitative experiments, we take three object classes as examples: cars, sofas, and chairs.
\autoref{table:TrainingDetail} shows details of the training datasets.
Each dataset consists of both real images and normal maps.
Real images were sampled from the LSUN dataset~\cite{Yu2015} with a simple filtering process to select images showing a single and sufficiently large target object.
Using a pre-trained object detector~\cite{Huang2017b}, we accepted images with only one bounding box of the target class whose area is larger than 25\% of the whole image.
After the filtering process, there were in total 83,765, 151,758, and 386,370 images for sofa, chair, and car, respectively.
These images were extended to 1:1 aspect ration by filling the borders by zero padding.
Figures~\ref{fig:TrainingSamples} (a), (b), and (c) show samples of the sofa, chair, and car images used for training.
The top row is real images from the LSUN dataset after post-processing, and the bottom row is normal maps rendered using models from the ShapeNet dataset.
As can be seen in the cases such as the top-middle example in Figs.~\ref{fig:TrainingSamples} (b) and the top-left example in (c), the real images still contain some occlusions and mismatched object poses compared to the normal maps even after the automatic filtering, which illustrates the fundamental difficulty of handling unpaired data.

We used 3D models taken from the ShapeNet dataset~\cite{Chang2015} to render normal maps.
Using 3,173, 6,778, and 3,385 models for sofa, chair, and car, the normal maps were rendered so that the pose distribution roughly resembles the real image dataset.
\autoref{table:TrainingDetail} lists the ranges of camera poses for each object, where the virtual camera was placed with increments of 5 degrees.
In total, there were 114,228, 515,128, and 257,260 normal maps for sofa, chair, and car.
Since the position of the object also differs in the real images, during training we also applied random shifting and scaling to these normal maps.

\begin{table}[t]
\centering
\caption{Training detail about the number of real data, 3D model, normal map, and the range of camera angles. When azimuth and elevation are 0, it means the camera is located in front of the object. The camera moves in increments of 5 degrees.}
\label{table:TrainingDetail}
\begin{tabular}{|l|l|l|l|l|l|l|}
\hline
\multirow{2}{*}{} & \multicolumn{1}{c|}{\multirow{2}{*}{Num. images}} & \multicolumn{1}{c|}{\multirow{2}{*}{Num. 3D models}} & \multicolumn{3}{c|}{Camera angle {[}degrees{]}} & \multicolumn{1}{c|}{\multirow{2}{*}{Num. normal maps}} \\ \cline{4-6}
 & \multicolumn{1}{c|}{} & \multicolumn{1}{c|}{} & \multicolumn{1}{c|}{Azimuth} & \multicolumn{1}{c|}{Elevation} & \multicolumn{1}{c|}{Num. angles} & \multicolumn{1}{c|}{} \\ \hline
Sofa & 83,765 & 3,173 & -45$\sim$45& 10$\sim$25& 36 & 114,228 \\ \hline
Car & 386,370 & 3,385 & -90$\sim$90 & 0$\sim$15 & 76 & 257,260 \\ \hline
Chair & 151,758 & 6,778 & -90$\sim$90 & 10$\sim$25 & 76 & 515,128 \\ \hline
\end{tabular}
\end{table}

\subsubsection{Baseline Methods}\label{Baseline}
Although there is no other method directly addressing the same task of shape-conditioned image generation, we picked two closely related approaches as baseline methods: SimGAN~\cite{Shrivastava2016} and pixel-level domain adaptation (PixelDA) network~\cite{Silberman}.
The network architectures, discriminator losses, and training hyper-parameters of these baseline methods were set to the same as our method (\modelname) for fair comparison, while method-specific losses stayed the same as the original papers.
Following the original method, SimGAN does not have the input appearance vector and there is no mechanism to change the appearance of generated images.
Since these methods were designed to modify rendered images of textured 3D models, we also evaluated them using textured rendering as input condition.
The textured images were rendered in the same settings as the normal maps.

\subsection{Comparison of Generated Images}

\begin{figure}[t]
	\begin{center}
	\includegraphics[width=\linewidth]{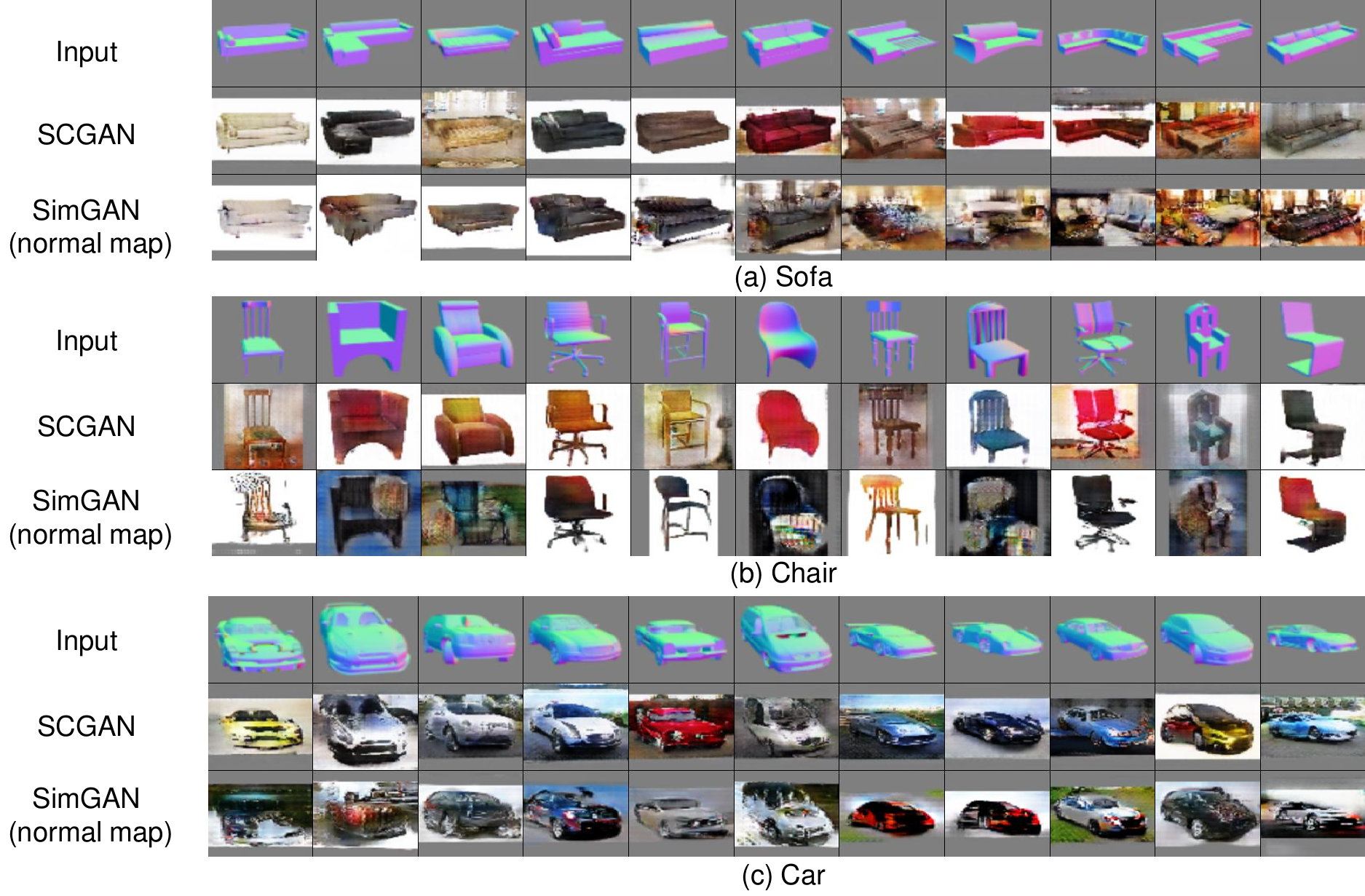}
	\end{center}
	\caption{Comparison with SimGAN and PixelDA for each object class. In each figure, the first row is input normal maps. The second and third rows are output from \modelname~and SimGAN using these normal maps.
	}
	\label{fig:RWComparison}
\end{figure}

\autoref{fig:RWComparison} shows examples of generated images from each method.
Figures~\ref{fig:RWComparison} (a), (b), and (c) correspond to the cases of sofa, chair, and car, respectively.
In each figure, the first row shows the input normal maps, and the second and third rows show the output from \modelname~and SimGAN using these normal maps as input.

It can be seen that \modelname~generates more naturalistic images than baseline methods.
SimGAN could not successfully modify normal maps and failed to generate realistic images in most cases.
In addition, there are many cases where the baseline methods failed to generate realistic background in Fig.~\ref{fig:RWComparison} (a).
This illustrates the advantage of our method which does not rely on a strong constraint unlike baseline methods minimizing the distance between the generated and input images.
\begin{figure}[t]
	\begin{center}
	\includegraphics[width=\linewidth]{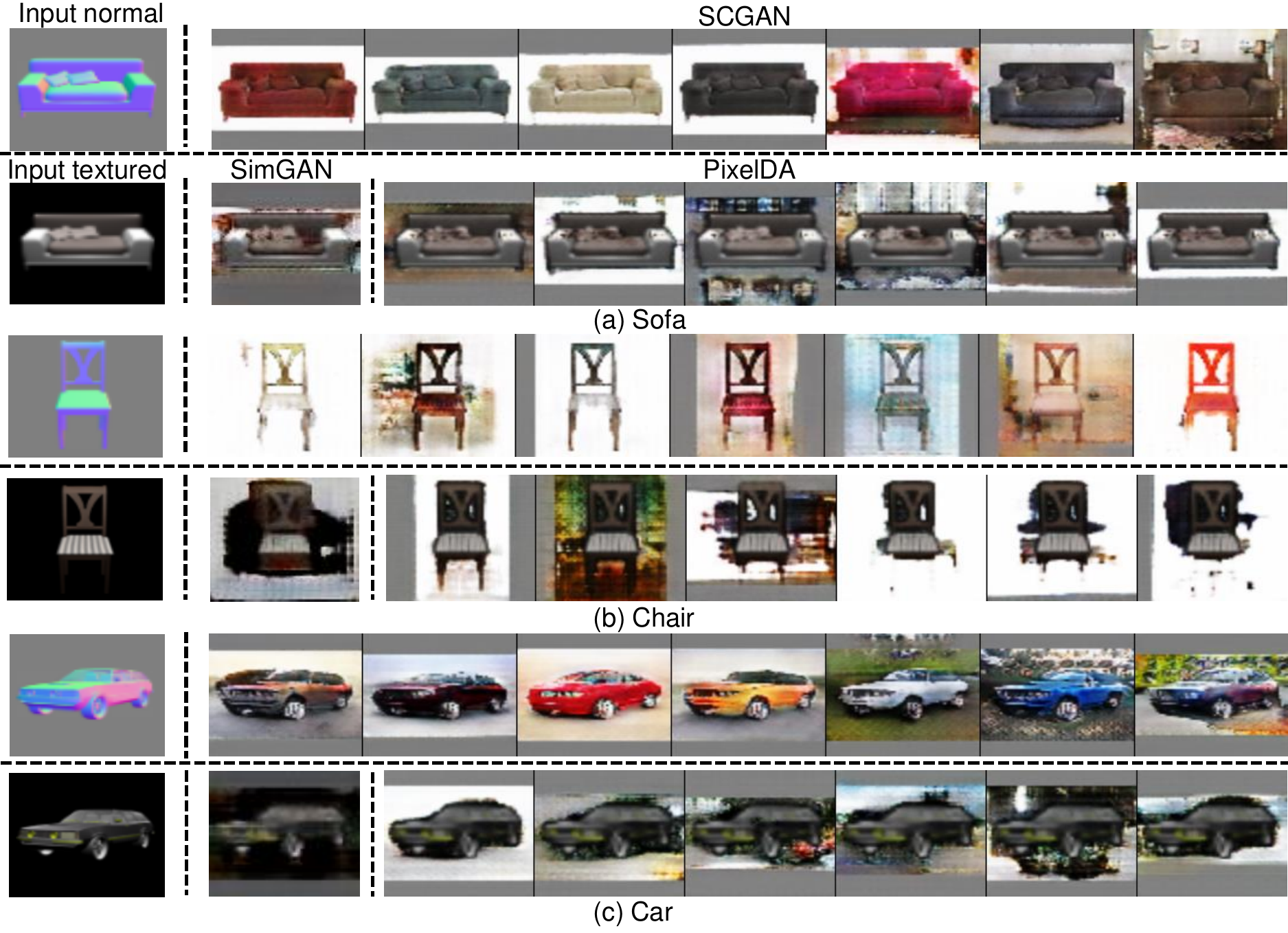}
	\end{center}
	\caption{Examples of generated images using the same normal map with several different appearance vectors. Each image shows the input normal map and textured image in the first column, the rest shows generated images with \modelname, SimGAN, and PixelDA.}
	\label{fig:OneNormalMap}
\end{figure}

\autoref{fig:OneNormalMap} further shows more output examples of \modelname, using the same normal map but with different appearance vectors.
Figures~\ref{fig:OneNormalMap} (a), (b), and (c) are the input normal map and generated images of sofa, chair, and car, respectively.
In each figure, the first column shows the input normal map and textured image.
The remaining first row shows the generated images from the normal map using \modelname, and the second row shows the output images from the textured image using both SimGAN and PixelDA.
Since SimGAN cannot control the output image appearance, it only shows one example.
While the baseline methods cannot control object shapes separately from the appearance, \modelname~can generate images with the same shape and diverse appearances.
It is noteworthy that in Fig.~\ref{fig:OneNormalMap} (a) the output images also keep the cushion placed on the sofa, which is not an easy case for keypoint-based methods.

\subsubsection{Ablation Study}

\begin{figure}[t]
	\begin{center}

    \includegraphics[width=\linewidth]{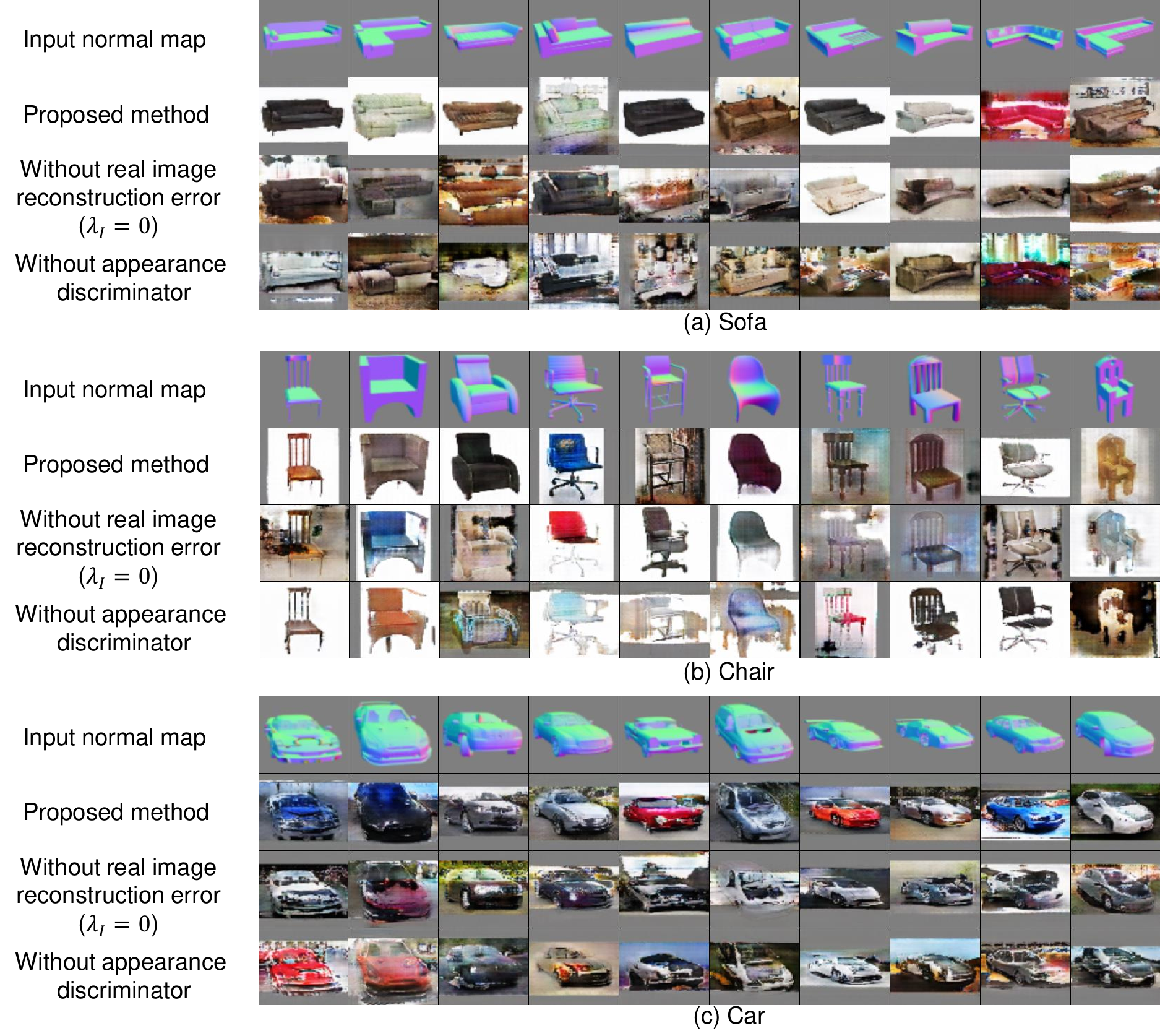}
	\end{center}
	\caption{Generated images without real image reconstruction error and appearance discriminator. The first rows show input normal maps, and the rest shows output images generated by \modelname, without the real image reconstruction loss, without the appearance discriminator loss.}
	\label{fig:PMComparison}
\end{figure}

In Fig.~\ref{fig:PMComparison}, we further show the effectiveness of individual loss terms in Eq.~(\ref{CycleWGANLoss}).
To demonstrate the effect of the proposed architecture using the separate appearance modeling and the cycle-consistent real image reconstruction loss, we evaluated models trained without real image reconstruction error and appearance discriminator.
Figures~\ref{fig:PMComparison} (a), (b), and (c) correspond to the cases of sofa, chair, and car, respectively.
In each figure, the first row shows the input normal maps.
The second row shows the output using all losses in Eq.~(\ref{CycleWGANLoss}).
The third row corresponds to the training result without the real image reconstruction error ($\lambda_I$ was set to zero), and the fourth row corresponds to the case trained without the appearance discriminator.

These examples show that the proposed approach improves the overall image quality by using these losses.
The real image reconstruction error significantly contributes to the realism of generated images, and the results without image reconstruction error mostly failed to generate object appearances.
When the network was trained without the appearance discriminator, the generated images sometimes become highly distorted as can be seen in middle columns of Fig.~\ref{fig:PMComparison} (a).

\subsubsection{Appearance Representation}
\begin{figure}[t]
	\begin{center}
    \includegraphics[width=\linewidth]{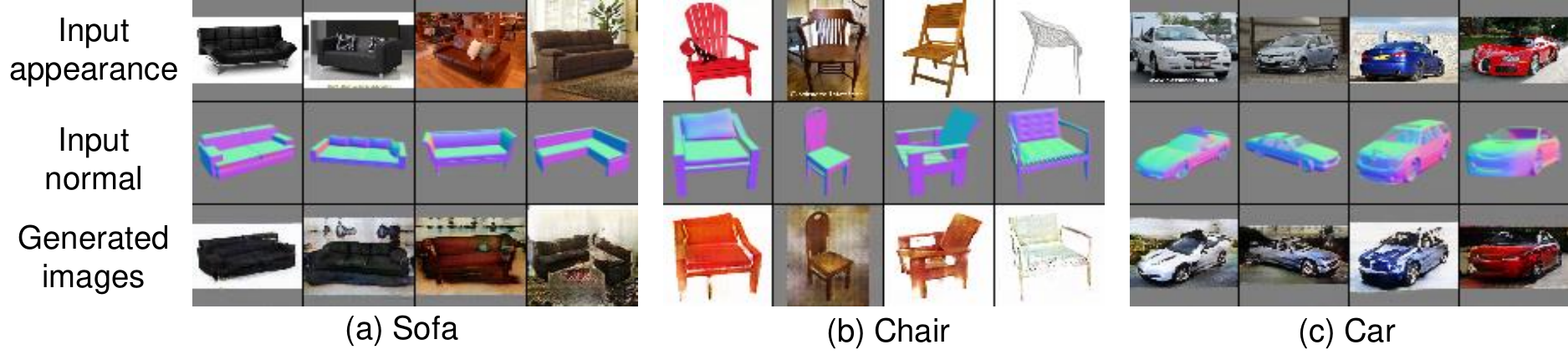}
	\end{center}
	\caption{Generated images with sampled appearances from real images. The first and second rows are input source images for appearance vectors and normal maps, and the third row shows the generated images.}
	\label{fig:NoiseFromReal}
\end{figure}

As a consequence of the cycle-consistent training, the appearance generator $G_z$ can also be used to extract appearance vectors from real images for generating new images.
\autoref{fig:NoiseFromReal} shows some examples of images generated using appearances sampled from real images.
Figures~\ref{fig:NoiseFromReal} (a), (b), and (c) correspond to the cases of sofa, chair, and car, respectively.
As can be seen in these examples, \modelname~can generate shape-conditioned images with the similar appearance with the source images.
This illustrates the potential of \modelname~for modifying pose and shape of objects in existing images.

\subsubsection{Handling Unknown 3D Shapes}

\begin{figure}[t]
	\begin{center}
    \includegraphics[width=\linewidth]{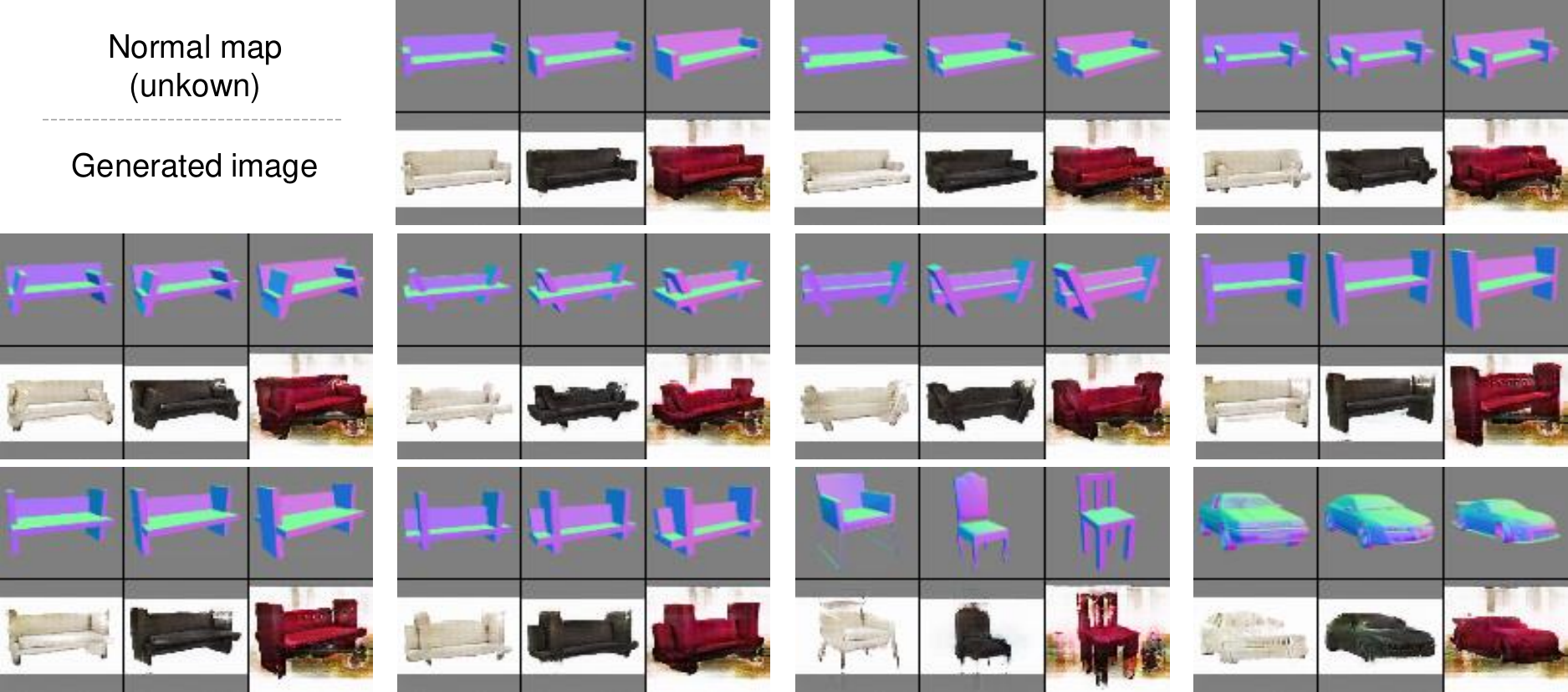}
  	\end{center}
\caption{Examples of generated images from unknown 3D shapes. Each block corresponds to one 3D model.
The first rows are input normal maps, and the second rows are generated images.}
\label{fig:RoughModelTestSofa}
\end{figure}

Another advantage of our method is that it can take an arbitrary normal map as input, even ones rendered using hand-crafted objects.
In Fig.~\ref{fig:RoughModelTestSofa}, we further show the output from the sofa image generator using hand-crafted sofa objects and shapes from the other object classes.
The hand-crafted models were created by a person who has never experienced 3D modeling, and consists of basic 3D shapes without any texture.

Each block corresponds to the result of one 3D model, with the same three appearances.
The first rows are input normal maps, and the second rows are generated images.
Even when the object shape is significantly different from ordinary sofa shapes, \modelname~successfully generates their corresponding images.
As can be seen in the bottom-right blocks, the proposed method tries to map the object texture to the input shape even when the shape comes from completely different object categories.

\subsection{Training Data Generation for Object Pose Estimation}

Since \modelname~also keeps the object pose as the same as the input normal maps, the generated images can serve as a training data for appearance-based object pose estimation.
In this section, we compare the effectiveness of \modelname~as training data generation framework, by comparing the accuracy of trained pose estimator with the cases using generated images from baseline methods.
The architecture of the pose estimation network follows the DenseNet~\cite{Huang2017}, while the last fully connected layer is modified to output 3-dimensional pose parameters.
The network weights were pre-trained on the ImageNet~\cite{deng2009imagenet}, and the whole network including the last layer was trained on each target object dataset.
Object poses are represented as Euler angles (azimuth, elevation, theta), and the loss function is set to be the Euclidean distance between ground-truth and estimated poses.

Test data were taken from the ObjectNet3D dataset~\cite{Xiang2016} which consists of images annotated with pose-aligned 3D models.
We selected images with the corresponding object annotations, and whose object poses stay within the pose range set for the training data.
In total, we used 886, 2,547, and 3,939 test images for sofas, chairs, and cars, respectively.

We compare the performance of the pose estimator with the ones trained using data generated by SimGAN~\cite{Shrivastava2016} and PixelDA~\cite{Silberman}.
As in the training of image generators, random shifting and scaling were also applied to the normal maps.
As an indicator of the upper-bound accuracy of this task, we also trained the same pose estimator using the test data via 5-fold cross-validation.
In addition, we evaluated the pose estimator directly using the textured images to show the estimator performance without any domain adaptation.
Similarly, to show the baseline performance of each task we also evaluated a na\"ive estimator which always output the mean pose in each object category.

\autoref{table:PoseComparison} lists pose estimation errors for each method and object category.
The estimation error was evaluated as the geodesic distance between the ground-truth rotation matrix $\bm R_t$ and the estimated rotation matrix $\bm R$ as $\frac{1}{\sqrt{2}}||\log{(\bm R^T \bm R_t)}||$~\cite{Xiang2016}.
The first column ({\em Target data}) shows the upper-bound performance obtained via cross-validation.
The second and third columns show the result using the dataset generated from normal maps, with \modelname~and SimGAN, respectively.
Similarly, the third and fourth columns show the result using the dataset generated from textured images, with SimGAN and PixelDA, respectively.
The fifth column ({\em No op.}) additionally shows the result directly using the original textured images.
The sixth column shows the na\"ive baseline performance of the average predictor.

The result shows that \modelname~achieved better pose estimation performances than SimGAN-based training results using normal maps, and better or close performance in comparison with SimGAN and PixelDA based training using textured 3D model images.
\modelname~significantly improved the pose estimation performance especially in the case of the chair dataset.
This is mainly because chair images have larger appearance gaps from the textured images, and \modelname~successfully generated training images closer to the actual test images.

\begin{table}[t]
\centering
\caption{Mean pose error for ObjectNet3D dataset when the pose estimator is trained using dataset generated by SimGAN, PixelDA, and textured images. Na\"ive baseline means that all predictions are an average of the target data.}
\label{table:PoseComparison}
\begin{tabular}{|l|l|l|l|l|l|l|l|}
\hline
\multirow{2}{*}{} & \multicolumn{1}{c|}{\multirow{2}{*}{Target data}} & \multicolumn{2}{c|}{Normal map} & \multicolumn{3}{c|}{Textured} & \multicolumn{1}{c|}{\multirow{2}{*}{Na\"ive baseline}} \\ \cline{3-7}
 & \multicolumn{1}{c|}{} & \multicolumn{1}{c|}{\modelname} & \multicolumn{1}{c|}{SimGAN~\cite{Shrivastava2016}} & SimGAN~\cite{Shrivastava2016} & \multicolumn{1}{c|}{PixelDA~\cite{Silberman}} & \multicolumn{1}{c|}{No op.} & \multicolumn{1}{c|}{} \\ \hline
Sofa & 21.1 & \textbf{23.9} & 27.1 & \textbf{23.8} & 24.9 & 28.7 & 30.0 \\ \hline
Chair & 21.5 & \textbf{26.0} & 33.8 & 35.2 & 28.2 & 41.3 & 47.9 \\ \hline
Car & 16.9 & \textbf{18.2} & 26.0 & 22.4 & 20.4 & 33.6 & 38.9 \\ \hline
\end{tabular}
\end{table}

%% file: conclusion.tex

\section{Conclusion}

In this work, we proposed \modelname, a GAN architecture for shape-conditioned image generation.
Given a normal map of the target object category, \modelname~generates images with the same shape as the input normal map.
The network can be trained without relying on paired training data with cycle-consistency losses, and it is able to generate images with diverse appearances through the latent modeling of image appearances.
Unlike prior work on conditional image generation, our method does not rely on any object-specific keypoint design and can handle arbitrary object categories.
The proposed method therefore provides a flexible and generic framework for shape-conditioned image generation tasks.

We demonstrated the advantage of \modelname~through both qualitative and quantitative evaluations.
\modelname~not only improves the quality of generated images while maintaining the input shape, but also efficiently handles the training data synthesis task for appearance-based object pose estimation.
In future work, we will further investigate applications of the proposed method including a wider range of learning-by-synthesis approaches, together with more detailed human evaluation on generated images.